\documentclass[letterpaper, 10 pt, conference]{ieeeconf}  % Comment this line out if you need a4paper

\IEEEoverridecommandlockouts                              % This command is only needed if 
\usepackage[noadjust]{cite} % TODO: Make sure if actually allowed
\usepackage{graphics} % for pdf, bitmapped graphics files
\usepackage{graphicx}
\usepackage{epsfig} % for postscript graphics files
\usepackage{mathptmx} % assumes new font selection scheme installed
\usepackage{booktabs}
\usepackage{tabularx}
\usepackage{amsmath} % assumes amsmath package installed
\usepackage{amssymb}  % assumes amsmath package installed
\usepackage{caption}
\usepackage{subcaption}
\usepackage{multirow}
\usepackage{tcolorbox}
\tcbuselibrary{breakable}

\title{\LARGE \bf
Vision Foundation Model Embedding-Based Semantic Anomaly Detection}

\author{Max Peter Ronecker$^{1,2}$, Matthew Foutter$^{3}$, Amine Elhafsi$^{3}$, Daniele Gammelli$^{3}$,  Ihor Barakaiev$^{3}$, \\Marco Pavone$^{3,5}$  and Daniel Watzenig$^{2,4}$% <-this % stops a space
\thanks{$^{1}$SETLabs Research GmbH, 80687 Munich, Germany
        {\tt\small first.last@setlabs.de}}%
\thanks{$^{2}$ Graz University of Technology, Institute of Visual Computing, 8010 Graz, Austria
        {\tt\small first.last@tugraz.at}}%
\thanks{$^{3}$ Stanford University, Autonomous Systems Laboratory, Stanford, CA, USA
        {\tt\small [mfoutter, amine, gamelli, igorb, pavone]@stanford.edu}}%
\thanks{$^{4}$ Virtual Vehicle Research GmbH, 8010 Graz, Austria
        {\tt\small first.last@v2c2.at}}%
\thanks{$^{5}$ NVIDIA, Santa Clara, USA
        {\tt\small pavone@nvidia.com}}%
}

\begin{document}

\maketitle
\thispagestyle{empty}
\pagestyle{empty}

%%%%%%%%%%%%%%%%%%%%%%%%%%%%%%%%%%%%%%%%%%%%%%%%%%%%%%%%%%%%%%%%%%%%%%%%%%%%%%%%
\begin{abstract}
Semantic anomalies are contextually invalid or unusual combinations of familiar visual elements that can cause undefined behavior and failures in system-level reasoning for autonomous systems. This work explores semantic anomaly detection by leveraging the semantic priors of state-of-the-art vision foundation models, operating directly on the image. We propose a framework that compares local vision embeddings from runtime images to a database of nominal scenarios in which the autonomous system is deemed safe and performant. In this work, we consider two variants of the proposed framework: one using raw grid-based embeddings, and another leveraging instance segmentation for object-centric representations. To further improve robustness, we introduce a simple filtering mechanism to suppress false positives. Our evaluations on CARLA-simulated anomalies show that the instance-based method with filtering achieves performance comparable to GPT-4o, while providing precise anomaly localization. These results highlight the potential utility of vision embeddings from foundation models for real-time anomaly detection in autonomous systems.
\end{abstract}

%%%%%%%%%%%%%%%%%%%%%%%%%%%%%%%%%%%%%%%%%%%%%%%%%%%%%%%%%%%%%%%%%%%%%%%%%%%%%%%%
\section{INTRODUCTION}
Autonomous vehicles, such as Waymo~\cite{waymo2024scaling} or Tesla~\cite{nedelea2022fsd}, are increasingly deployed in real-world environments and rely heavily on machine learning (ML) algorithms, especially in perception modules, e.g., object detection. While these algorithms often perform reliably within their training distributions, ML models remain vulnerable to out-of-distribution (OOD) inputs, which can lead to unsafe or unpredictable behavior. An OOD input is data that significantly differs from the training distribution of an ML model and is defined relative to that model, such as unusual objects, rare weather conditions, or novel environments. A common mitigation strategy to avoid unpredictable failure modes involves detecting OOD situations at runtime and transitioning the system to a safe state~\cite{SinhaSharmaEtAl2022}.

\begin{figure}[t!]
  \centering
  \includegraphics[width=\linewidth]{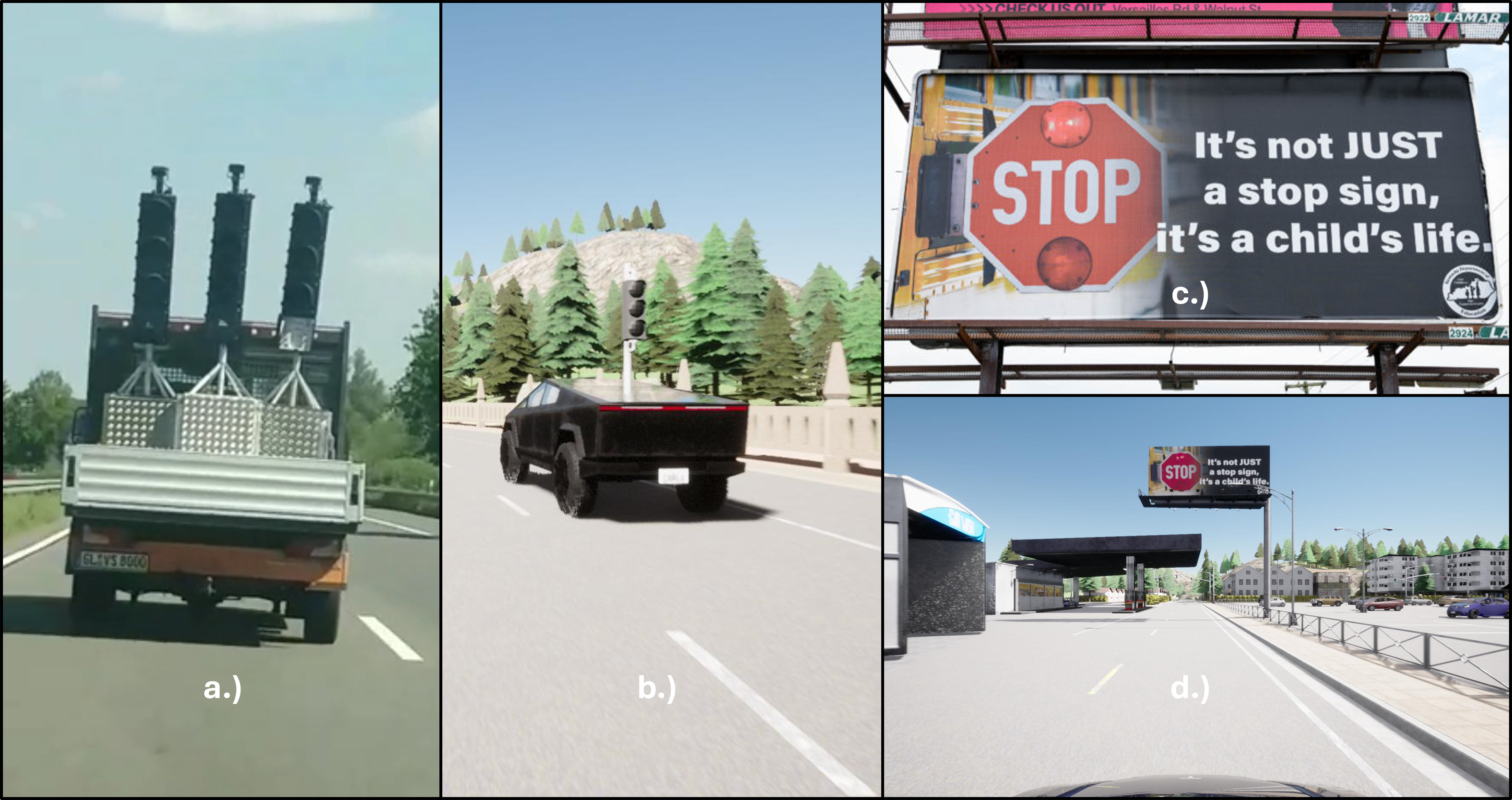}
\caption{Examples of semantic anomalies (a,c) and their CARLA-simulated equivalents (b,d). A truck with traffic lights (a) confused a Tesla into detecting active signals \cite{robitzski2021trafficlight}. A stop sign on a billboard (c) caused unintended braking \cite{kelly2017stopSign,robitzski2021stopsign}.}
  \label{fig:sem_an_examples}
\vspace{-3mm}
\end{figure}

Among OOD observations, semantic anomalies pose a unique challenge. As defined in~\cite{ElhafsiSinhaEtAl2023}, semantic anomalies reflect failures in high-level reasoning rather than low-level perception or control. In contrast to traditional OOD cases, semantic anomalies are defined with respect to the system's context and operational domain and involve in-distribution elements arranged in atypical or contextually invalid ways---for example, a stop sign on a billboard or a traffic light mounted on a moving truck may confuse an autonomous vehicle, despite all visual components being familiar. These examples are inspired by real-life occurrences, such as those observed in the Tesla system~\cite{robitzski2021trafficlight,robitzski2021stopsign} shown in Fig.~\ref{fig:sem_an_examples}. 
These anomalies are often trivial for humans to interpret but difficult for conventional ML models to detect. The emergence of Large Language Models (LLMs) and Vision-Language Models (VLMs) provides new tools for addressing this gap, leveraging their strong reasoning and generalization capabilities for semantic anomaly detection using textual prompts, multi-modal embeddings, or direct image-based reasoning~\cite{ElhafsiSinhaEtAl2023,SinhaElhafsiEtAl2024}. 

While foundation models offer zero-shot capabilities~\cite{clip_2021,mae_2022,oquab2023dinov2} and perform well on nominal data, these high-capacity ML models often suffer from high latency—with response times of several seconds—and are prone to hallucination~\cite{hallucination_survey_2025}.
Embedding-based methods are faster but typically provide only a coarse anomaly score without spatial localization~\cite{SinhaElhafsiEtAl2024}. This motivates the development of efficient, embedding-based approaches that can both detect and localize semantic anomalies in real time. In this work, we take a first step toward this goal. 

We evaluate existing semantic anomaly detectors and propose a vision embedding-based framework for anomaly detection and localization. The framework is tested on simulated CARLA data, following~\cite{ElhafsiSinhaEtAl2023,SinhaElhafsiEtAl2024}. 

Our contributions are as follows:
\begin{itemize}
    \item \textit{Evaluation of existing VLM-based semantic anomaly detectors:} We provide a detailed evaluation of existing VLM-based semantic anomaly detectors at the frame level, identifying their strengths and weaknesses.
    
    \item \textit{Embedding-based semantic anomaly detection:} We propose two variations of a foundation model embedding-based semantic anomaly detection framework that operates directly on images. Our approach achieves detection performance comparable to large vision-language models such as GPT-4o, while also enabling precise localization of anomalies. Extensive evaluation on simulated data highlights the potential of vision embeddings for semantic anomaly detection.
    
    \item \textit{Filtering techniques for embedding-based semantic anomaly detection:} We introduce a simple yet effective filtering technique that further boosts the performance of our embedding-based framework, helping it reach GPT-4o-level results.
\end{itemize}

\section{RELATED WORK}
\subsection{Vision Foundation Models}

Foundation models are large-scale neural networks trained on broad, often internet-scale data to perform well across many tasks with minimal adaptation~\cite{bommasani2021a}. Models like CLIP~\cite{radford2021clip}, DINO~\cite{caron2021dino}, and DINOv2~\cite{oquab2023dinov2,darcet2023vitneedreg} have demonstrated the capability of large-scale pretraining for learning general-purpose visual features. CLIP learns joint image-text embeddings from 400 million image-caption pairs, enabling zero-shot classification across diverse categories. DINO introduced self-supervised learning using Vision Transformers (ViTs), showing that meaningful object-centric representations can emerge without labels. DINOv2 builds on this by training on 142 million curated images, yielding robust and transferable features across a wide range of visual tasks (e.g., image and instance recognition) with minimal fine-tuning.

Segment Anything~\cite{kirillov2023segmentanything} and its follow-up model~\cite{ravi2025sam} introduce general-purpose segmentation models trained on large-scale instance mask datasets. These models enable automatic or promptable segmentation of arbitrary objects, using object detectors as prompts~\cite{minderer2022owl,minderer2023scaling}. Combined, these models can potentially be used to provide semantically rich embeddings that can support anomaly detection by identifying distributional shifts or unexpected objects in the scene.

\subsection{Foundation Models for Anomaly Detection in Robotics}

Semantic anomalies, as defined in \cite{ElhafsiSinhaEtAl2023} refer to failures in high-level reasoning rather than low-level perception or control. Examples include autonomous vehicles braking for stop signs on billboards or failing to interpret unusual traffic configurations. A proposed approach leverages large language models (LLMs) to monitor a robot’s perceptions and decisions, flagging behaviors that deviate from human intuition. Experiments in driving and manipulation tasks show that such LLM-based monitors align well with human judgment.

Sinha et al. \cite{SinhaElhafsiEtAl2024} introduce a two-stage system: a lightweight classifier detects anomalies from LLM/VLM embeddings, triggering a slower generative LLM to reason and suggest recovery. Coupled with model-predictive control, this enables safe replanning. The fast detector outperforms GPT-4 in failure detection, showing embeddings suffice for real-time use.

Other foundation model-based approaches for anomaly and OOD detection include S2M \cite{sam2_OOD_2024}, which converts anomaly scores into segmentation masks by generating box prompts for the segmentation model SAM, and AnomalyCLIP \cite{zhou2023anomalyclip}, a zero-shot method that learns object-agnostic prompts and applies a glocal loss for accurate detection and segmentation without target domain data.

However, these methods either require large models and significant compute, making real-time processing difficult, fail to localize the anomaly, or are not suited for complex semantic anomalies. The proposed method aims to provide a lightweight, embedding-based approach capable of detecting and localizing semantic anomalies.

\section{METHODOLOGY}

\begin{figure}[t!]
  \centering
  \includegraphics[width=\linewidth]{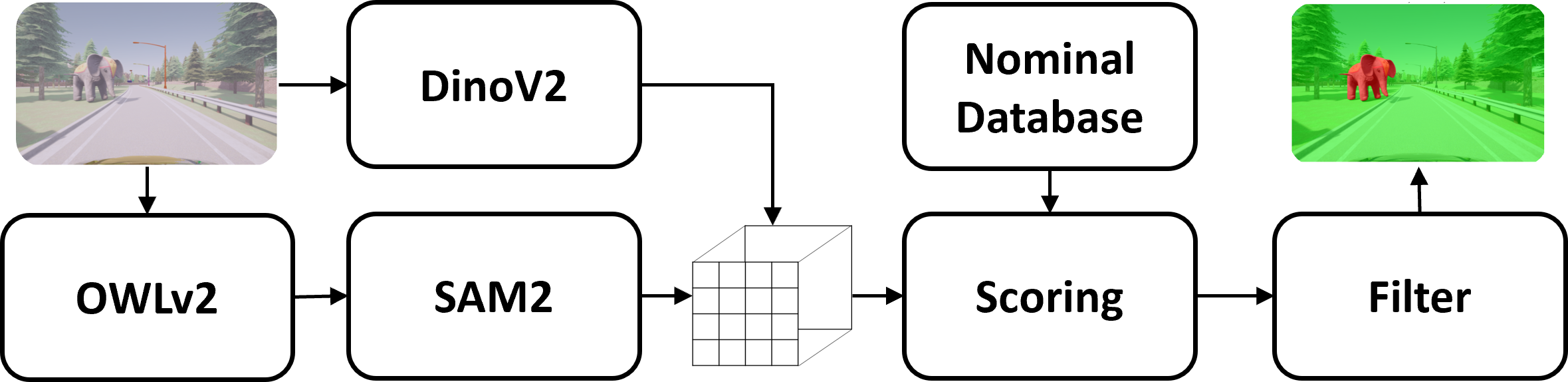}
  \caption{Overview of the proposed vision-based semantic anomaly detection framework. Structuring semantic anomaly detection in this way enables detecting anomalies without requiring access to out-of-distribution data.}
  \label{fig:system}
  \vspace{-3mm}
\end{figure}

\subsection{Problem Formulation}
\label{sec:problem-formulation}

This work addresses semantic anomaly detection and localization using vision foundation model embeddings. The core idea is that such embeddings encode meaningful semantic information, enabling compact representations of visual scenes. Following~\cite{SinhaElhafsiEtAl2024}, the approach assumes access to a database of embeddings from nominal scenarios previously encountered and successfully handled by the system. At runtime, embeddings of incoming images are compared to this database. If the distance to all known embeddings exceeds a defined threshold, the input is flagged as a semantic anomaly.

Evaluation is conducted using the dataset from~\cite{ElhafsiSinhaEtAl2023,SinhaElhafsiEtAl2024}, which contains CARLA-simulated autonomous driving scenes (CARLA version 0.9.15). The dataset includes nominal cases across multiple maps, semantic anomalies such as stop signs on billboards and trucks carrying traffic lights, and out-of-distribution objects like robots (see Fig.\ref{fig:sem_an_examples}). Each anomaly appears in multiple variations across different maps and positions. More examples are provided in Appendix~\ref{sec:example_images_scenarios}.

Anomaly detection performance is measured using frame-level binary classification metrics including F1-score, balanced accuracy, true positive rate (TPR), and false positive rate (FPR). For spatial localization, the dataset is extended with binary ground truth masks. A detection is considered a true positive (TP) if its Intersection over Union (IoU) with the ground truth mask is at least 0.3; otherwise, it is a false negative (FN). For anomaly-free frames, any predicted anomaly is counted as a false positive (FP).

\subsection{Proposed Approach}
The following components detail the embedding generation, anomaly detection, and filtering steps of the proposed system, as shown in Fig.~\ref{fig:system}.

\subsubsection{Embedding Generation}

The first step—either offline, for constructing the nominal database, or online, during inference, involves computing DINOv2 embeddings for the current image. DINOv2 produces 256 patch embeddings per image (number of patches \( p \)), each representing a \( 14 \times 14 \) pixel region, with an embedding dimension of \( d = 384 \).

Two variants of embedding extraction are considered. The first uses the standard grid-based patch embeddings directly from DINOv2. The second targets a more object-centric representation by combining OWLv2 and SAM2. OWLv2 generates prompts for object instances in the image, which are then segmented by SAM2. For each detected instance, the DINOv2 embeddings of all patches within the corresponding mask are averaged to create a single instance-level embedding. This procedure is applied to all instances in the image. Both approaches are evaluated and compared to determine their effectiveness for anomaly detection.

\subsubsection{Anomaly Detection}

Following~\cite{SinhaElhafsiEtAl2024}, anomaly detection is performed by comparing current observations to a set of nominal experiences. The nominal set consists of variable-length trajectories and their corresponding image observations \(\mathbf{o}_i\) that the autonomous vehicle can safely handle, and are therefore considered nominal. Instead of single image embeddings, DINOv2 patch-level embeddings are used for finer anomaly detection and localization (Fig.~\ref{fig:system}).

Each prior image \( \mathbf{o}_i \in \mathcal{D}_{\text{nom}} \) is embedded offline using DINOv2 (\( \phi(\cdot)\)), resulting in a cache of patch-level embeddings \( \mathcal{D}_e = \{\mathbf{e}_i\}_{i=1}^N \), where \( \mathbf{e}_i = \phi(\mathbf{o}_i) \in \mathbb{R}^{p \times d} \). \( N \) denotes the number of observations in the nominal set \( \mathcal{D}_{\text{nom}} \).

At runtime, a new observation \( \mathbf{o}_t \) is embedded as \( \mathbf{e}_t = \phi(\mathbf{o}_t) \), and an anomaly score \( s(\mathbf{e}_t; \mathcal{D}_e) \in \mathbb{R} \) is computed by comparing all patches in \( \mathbf{e}_t \) to the nearest patches in the nominal cache. A simple score function uses the maximum cosine similarity over all patch pairs, negated to represent dissimilarity:

\begin{equation}
s(\mathbf{e}_t; \mathcal{D}_e) := -\max_{\mathbf{e}_i \in \mathcal{D}_e} \max_{j,k} \frac{\mathbf{e}_t^{(j)^\top} \mathbf{e}_i^{(k)}}{\|\mathbf{e}_t^{(j)}\| \|\mathbf{e}_i^{(k)}\|} 
\label{eq:anomaly_score}
\end{equation}

An observation is classified as anomalous if its score exceeds a threshold \( \tau \). The threshold is estimated as the \( \alpha \)-quantile of anomaly scores computed over the nominal set in a leave-one-out fashion, where leave-one-out refers to excluding all embeddings from the same experiment. An experiment is defined as the full sequence of images collected under a specific configuration and map. This avoids nominal bias, as consecutive images are often very similar and could otherwise lead to a low anomaly threshold.

\subsubsection{Filtering}

Due to the small size of the nominal database and the tendency of instance segmentation to oversegment, small isolated patches with high anomaly scores may appear. These often lack semantic relevance and result in false positives. A simple post-processing step removes small connected components from the binary anomaly map based on a pixel threshold. This effectively reduces noise without requiring changes to the anomaly scoring. More advanced filtering methods are not explored in this work.

\section{EVALUATION AND DISCUSSION}

The evaluation compares the performance of the visual embedding-based anomaly detector against large language and vision-language models used in~\cite{ElhafsiSinhaEtAl2023,SinhaElhafsiEtAl2024}. Three systems are considered: a GPT-4o baseline (Version: gpt-4o-2024-11-20), an embedding-based method without instance information, and an instance embedding-based method. Thresholds for anomaly score and patch size are determined empirically. Details on evaluation metrics are provided in Section~\ref{sec:problem-formulation}.

\subsection{Result Analysis}
\label{sec:result_analysis}

\begin{figure*}[t]
    \centering
    % Subfigure: Embedding-based
    \begin{subfigure}[t]{0.49\textwidth}
        \centering
        \includegraphics[width=0.49\linewidth]{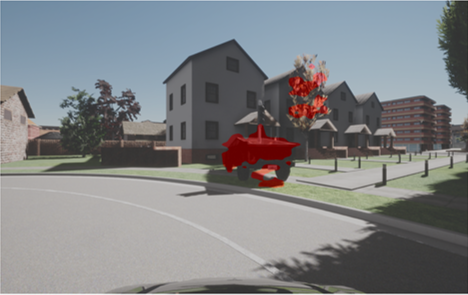}
        \includegraphics[width=0.49\linewidth]{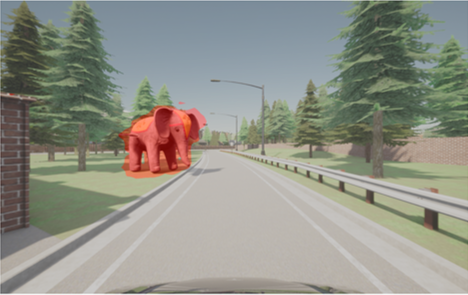}\\[0.2em]
        \includegraphics[width=0.49\linewidth]{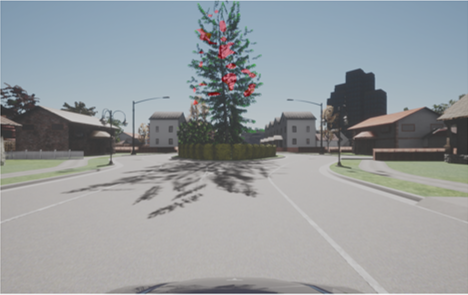}
        \includegraphics[width=0.49\linewidth]{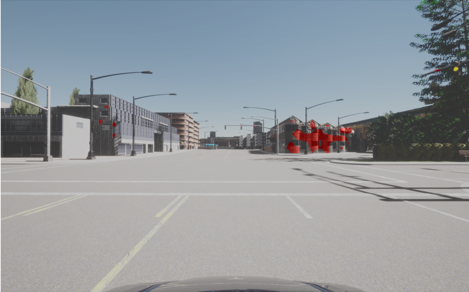}
        \caption{Embedding-based detection (top: TPs, bottom: FPs)}
    \end{subfigure}
    \hfill
    % Subfigure: Instance-based
    \begin{subfigure}[t]{0.49\textwidth}
        \centering
        \includegraphics[width=0.49\linewidth]{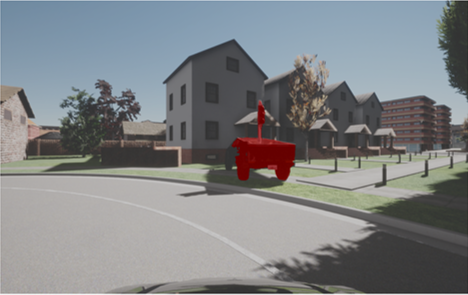}
        \includegraphics[width=0.49\linewidth]{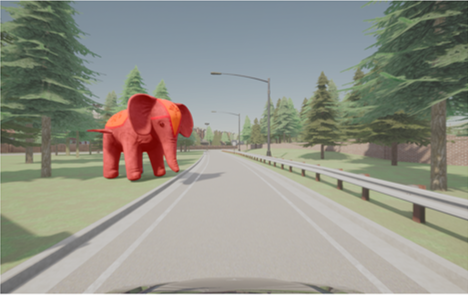}\\[0.2em]
        \includegraphics[width=0.49\linewidth]{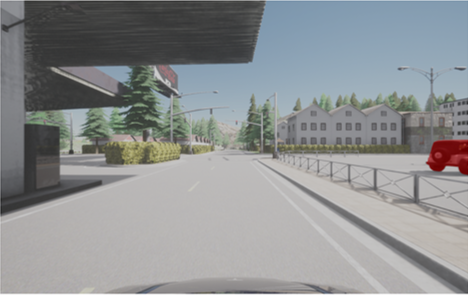}
        \includegraphics[width=0.49\linewidth]{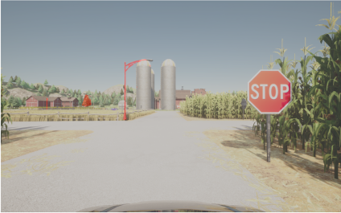}
        \caption{Instance-based detection (top: TPs, bottom: FPs)}
    \end{subfigure}

    \caption{Qualitative comparison of anomaly detections. Each method (embedding-based left, instance-based right) shows two true positives (top row) and two false positives (bottom row).}
    \label{fig:embedding_instance_grid}
    \vspace{-6mm}
\end{figure*}

As shown in Table~\ref{tab:combined_metrics}, GPT-4o achieves the highest F1-score and lowest FPR for OOD-objects. It performs well on obvious cases (e.g., an elephant or robot on the road) but often misclassifies more subtle semantic anomalies, such as traffic lights on trucks or stop signs on billboards as normal, resulting in low TPR for these scenarios. Examples of this are provided in Appendix~\ref{sec:gpt4o-failures}. 
This limitation likely stems from the zero-shot nature and the limited examples in the prompt (Section~\ref{sec:gpt4_prompt}), which may not sufficiently define semantic anomalies. Incorporating more anomaly examples through in-context learning or fine-tuning could improve performance. However, this introduces risks of overfitting and undermining the generalization benefits of foundation models. It also remains infeasible to represent the full range of possible anomalies. Additionally, domain shift between real-world training data and synthetic CARLA images may further impact performance in the tested scenarios. Overall, GPT-4o is strong at identifying nominal scenes, resulting in consistently low FPR. 
To provide a better understanding of performance and failure modes, Fig.~\ref{fig:embedding_instance_grid} shows exemplary true positives and false positives. These results were selected as representative examples of the overall behavior.

The embedding-based method without filtering is sensitive (high TPR) to OOD-objects but suffers from a high FPR across all scenarios, limiting overall performance. Its strong results on OOD-objects can be attributed to their absence in the nominal dataset, unlike elements such as billboards or traffic lights, which also appear in normal scenes. This suggests that the system primarily responds to visual novelty rather than true semantic anomalies, which often involve familiar objects in unusual combinations. This is further supported by cases where only a visually novel component (e.g., a Cybertruck, which is not present in the nominal dataset) is detected, rather than the full anomaly consisting of both the traffic light and the truck (see Fig.~\ref{fig:classification} (a), TPs). The same sensitivity to visual novelty likely contributes to the high FPR. False positives often arise from uncommon elements like vegetation patches or unusual buildings (Fig.~\ref{fig:classification} (a), FPs), which may be underrepresented in the nominal dataset. Expanding the nominal dataset could help mitigate this. Additionally, applying filtering techniques effectively reduces FPR and improves F1-score by removing isolated false detections caused by noisy patch-level scoring.

The instance-based method performs best in semantic anomaly scenarios and offers a more balanced trade-off between sensitivity and precision. With filtering, FPR is further reduced, and F1-score improves across all cases. It is particularly effective in the Stop Sign scenario, where it is the only embedding-based variant to achieve meaningful performance. On the full dataset, the instance-based method with filtering matches GPT-4o and outperforms it in semantic anomaly detection. It also produces sharper and more complete detections due to the use of instance segmentation masks, detecting full anomalies (e.g., traffic light and truck) that are often missed or partially detected by the patch-embedding-based method (see Fig.~\ref{fig:embedding_instance_grid} (a,b), TPs). However, it still suffers from false positives, though to a lesser extent than the patch-embedding-based method. Averaging scores within instance masks likely reduces the impact of outliers. Due to its object-centric design, the method often labels entire objects (e.g., a streetlight) as false positives (see Fig.~\ref{fig:embedding_instance_grid}b, FP). These errors are likely caused by unseen visual patterns and the limited semantic expressiveness of the embeddings. Another issue that limits performance is that the object detector used for instance segmentation occasionally struggles with synthetic CARLA images, leading over-segmented objects which in turn can cause anomalous fragments. Examples of the false positive detections and their respective segmentation masks are shown in Appendix ~\ref{sec:object_detector_segmentation}.

\begin{table}[t]
\centering
\caption{Overall and scenario-wise evaluation. NF = No Filter, F = Filter. Values in brackets show change from NF to F. Frame-level binary classification metrics are reported for the full dataset and individual scenarios.}
\label{tab:combined_metrics}
\begin{tabular}{lccc}
\toprule
\multicolumn{4}{c}{\textbf{Full Dataset}} \\
\midrule
\textbf{Method} & \textbf{TPR} & \textbf{FPR} & \textbf{F1} \\
\midrule
GPT-4o                  & 0.33 & \textbf{0.03} & 0.47 \\
Embedding (NF)         & 0.36 & 0.49 & 0.32 \\
Embedding (F)          & 0.36 (+0.00) & 0.36 (–0.13) & 0.36 (+0.04) \\
Instance (NF)          & 0.37 & 0.27 & 0.40 \\
Instance (F)           & \textbf{0.44} (+0.07) & 0.17 (–0.10) & \textbf{0.51} (+0.11) \\
\midrule
\multicolumn{4}{c}{\textbf{Traffic Light}} \\
\midrule
GPT-4o                  & 0.30 & \textbf{0.06} & 0.44 \\
Embedding (NF)         & 0.37 & 0.55 & 0.37 \\
Embedding (F)          & 0.37 (+0.00) & 0.38 (–0.17) & 0.40 (+0.03) \\
Instance (NF)          & 0.47 & 0.27 & 0.52 \\
Instance (F)           & \textbf{0.54} (+0.07) & 0.17 (–0.10) & \textbf{0.62} (+0.10) \\
\midrule
\multicolumn{4}{c}{\textbf{Stop Sign}} \\
\midrule
GPT-4o                  & \textbf{0.12} & \textbf{0.03} & 0.19 \\
Embedding (NF)         & 0.04 & 0.41 & 0.03 \\
Embedding (F)          & 0.04 (+0.00) & 0.33 (–0.08) & 0.04 (+0.01) \\
Instance (NF)          & 0.06 & 0.28 & 0.06 \\
Instance (F)           & 0.17 (+0.11) & 0.11 (–0.17) & \textbf{0.23} (+0.17) \\
\midrule
\multicolumn{4}{c}{\textbf{OOD-Objects}} \\
\midrule
GPT-4o                  & \textbf{0.62} & \textbf{0.08} & \textbf{0.71} \\
Embedding (NF)         & 0.57 & 0.48 & 0.50 \\
Embedding (F)          & 0.57 (+0.00) & 0.33 (–0.15) & 0.56 (+0.06) \\
Instance (NF)          & 0.31 & 0.23 & 0.38 \\
Instance (F)           & 0.37 (+0.06) & 0.18 (–0.05) & 0.45 (+0.07) \\
\bottomrule
\end{tabular}
\vspace{-6mm}
\end{table}

\subsection{Score Analysis}

\begin{figure*}[h]
  \centering
  \includegraphics[width=\textwidth]{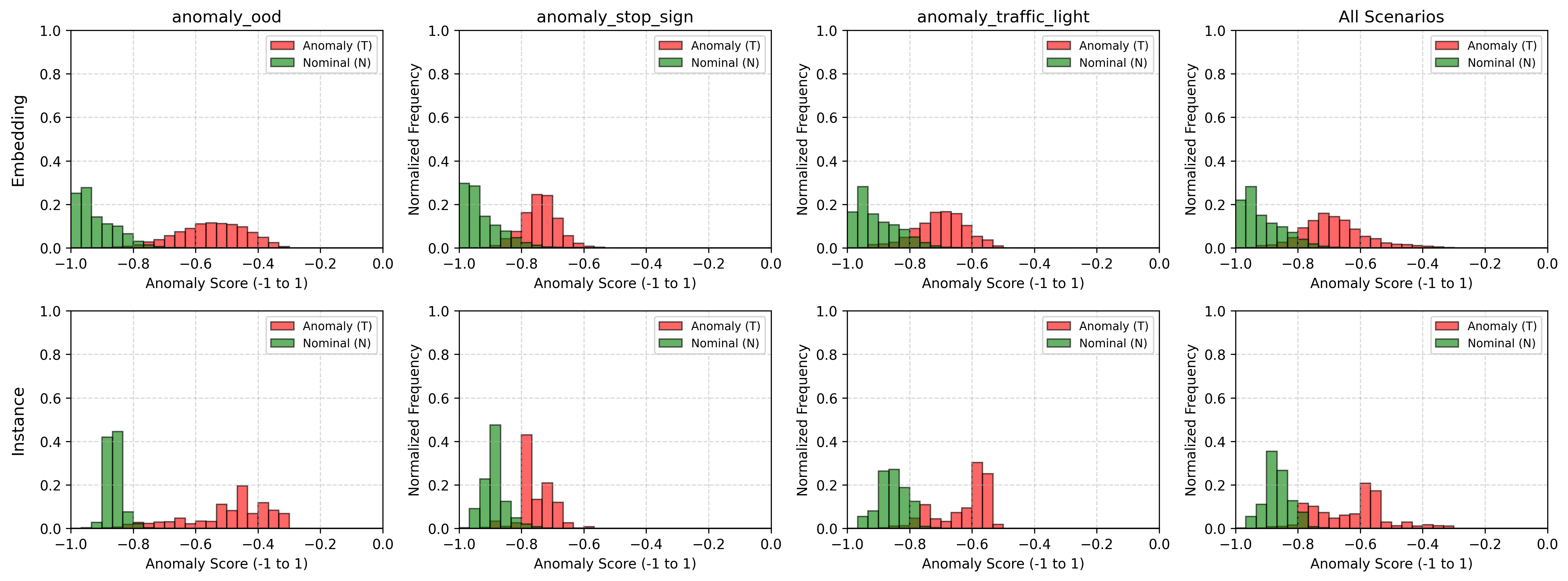}
  \caption{Distribution of anomaly scores for anomalies (T) and nominal objects (N) across scenarios.}
  \label{fig:score_distribution}
  %\vspace{-6mm}
\end{figure*}

\begin{figure*}[h]
  \centering
  \includegraphics[width=\textwidth]{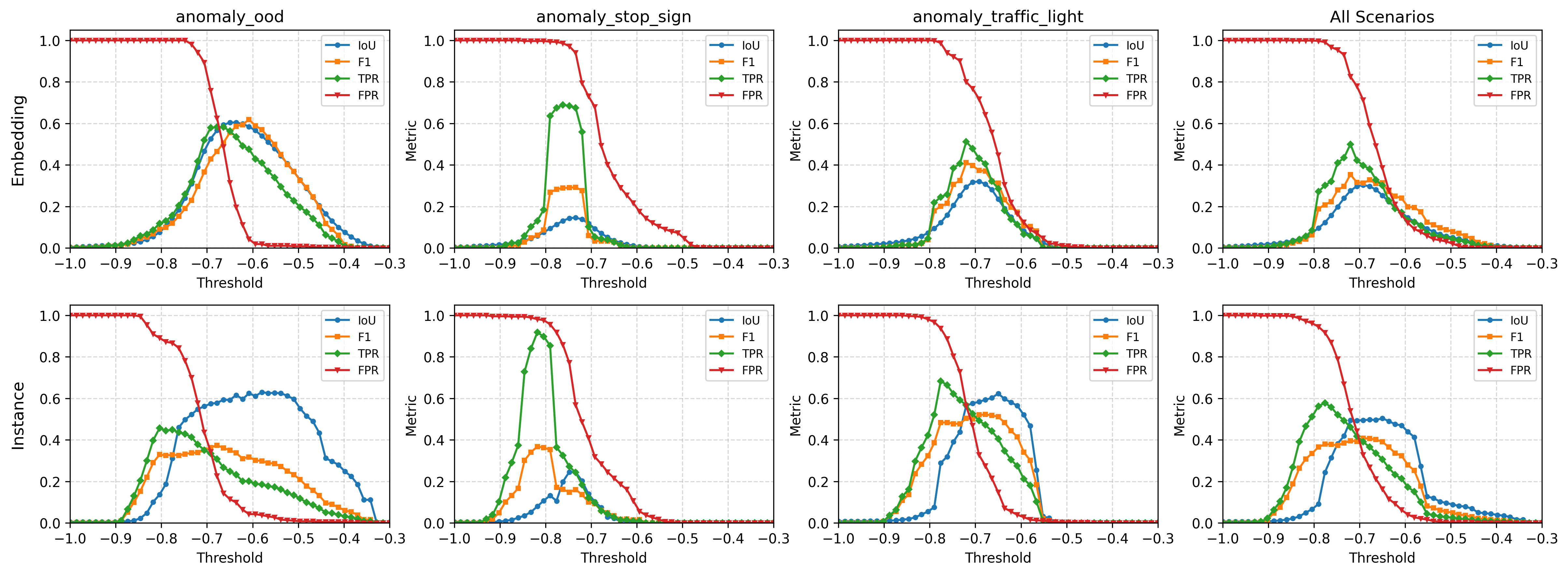}
  \caption{Threshold sweep showing metric trends (IoU, F1, TPR, FPR) for both methods across all scenarios.}
  \label{fig:score_metrics}
  %\vspace{-6mm}
\end{figure*}

Fig.~\ref{fig:score_distribution} shows the distribution of anomaly scores for both methods. In most cases, anomalies receive higher scores and nominal objects lower scores, resulting in an unexpectedly clear separation with minimal confusion across individual scenarios. Specifically, the instance-based method performs better in semantic anomaly scenarios, while the embedding-based method is more effective for OOD objects. This further indicates that vision foundation model embeddings have the potential to be used for OOD and semantic anomaly detection. 

However, clear separation is not always achieved. Different anomaly types yield different score ranges: OOD objects and the traffic light scenario produce the highest anomaly scores, while the stop sign scenario remains closer to nominal scores. This is consistent with previous results, where methods perform well in these scenarios but struggle with subtler cases like the stop sign. These varying score ranges complicate the selection of a single optimal threshold across all scenarios and indicate a misalignment of the different anomalies in the embedding space. Further analysis presented in Fig.~\ref{fig:score_metrics} confirms this observation. Higher true positive rates consistently come with more false positives, and the threshold for the best performance differs across scenarios. 

\section{CONCLUSION AND OUTLOOK}
This work takes a first step toward using vision foundation model embeddings for semantic anomaly and OOD detection. The results show that the proposed embedding-based approaches can detect and localize semantic anomalies, achieving performance comparable to GPT-4o. Overall, the presented framework provides a solid foundation for embedding-based semantic anomaly detection and motivates further research in this direction. To improve robustness, future work could focus on developing embeddings that better capture characteristics that constitute an anomaly or adopt more adaptive scoring mechanisms, such as energy-based models. In addition, analyzing embeddings in isolation might lead to a loss of global context, which is crucial for detecting anomalies resulting from atypical object combinations. Incorporating embeddings and scene structure into a graph-based representation may facilitate joint reasoning and better preserve contextual information.

\section*{ACKNOWLEDGEMENT}
This work has partially received funding from the German Federal Ministry for Economic Affairs and Climate Action (BMWK) under grant agreement 19I21039A.


%%%%%%%%%%%%%%%%%%%%%%%%%%%%%%%%%%%%%%%%%%%%%%%%%%%%%%%%%%%%%%%%%%%%%%%%%%%%%%%%

%%%%%%%%%%%%%%%%%%%%%%%%%%%%%%%%%%%%%%%%%%%%%%%%%%%%%%%%%%%%%%%%%%%%%%%%%%%%%%%%

%%%%%%%%%%%%%%%%%%%%%%%%%%%%%%%%%%%%%%%%%%%%%%%%%%%%%%%%%%%%%%%%%%%%%%%%%%%%%%%%

%%%%%%%%%%%%%%%%%%%%%%%%%%%%%%%%%%%%%%%%%%%%%%%%%%%%%%%%%%%%%%%%%%%%%%%%%%%%%%%%

\bibliographystyle{IEEEtran}
\bibliography{IEEEtranBST/IEEEabrv,IEEEtranBST/IEEEexample}

\begin{thebibliography}{10}
\providecommand{\url}[1]{#1}
\csname url@samestyle\endcsname
\providecommand{\newblock}{\relax}
\providecommand{\bibinfo}[2]{#2}
\providecommand{\BIBentrySTDinterwordspacing}{\spaceskip=0pt\relax}
\providecommand{\BIBentryALTinterwordstretchfactor}{4}
\providecommand{\BIBentryALTinterwordspacing}{\spaceskip=\fontdimen2\font plus
\BIBentryALTinterwordstretchfactor\fontdimen3\font minus \fontdimen4\font\relax}
\providecommand{\BIBforeignlanguage}[2]{{%
\expandafter\ifx\csname l@#1\endcsname\relax
\typeout{** WARNING: IEEEtran.bst: No hyphenation pattern has been}%
\typeout{** loaded for the language `#1'. Using the pattern for}%
\typeout{** the default language instead.}%
\else
\language=\csname l@#1\endcsname
\fi
#2}}
\providecommand{\BIBdecl}{\relax}
\BIBdecl

\bibitem{waymo2024scaling}
\BIBentryALTinterwordspacing
{Waymo}, ``Scaling waymo one safely across four cities this year,'' \url{https://waymo.com/blog/2024/03/scaling-waymo-one-safely-across-four-cities-this-year}, March 2024, accessed: 2025-04-07. [Online]. Available: \url{https://waymo.com/blog/2024/03/scaling-waymo-one-safely-across-four-cities-this-year}
\BIBentrySTDinterwordspacing

\bibitem{nedelea2022fsd}
\BIBentryALTinterwordspacing
A.~Nedelea, ``Any tesla driver can now join full self-driving beta regardless of safety score,'' \emph{InsideEVs}, November 2022, accessed: 2025-04-07. [Online]. Available: \url{https://insideevs.com/news/623469/tesla-fsd-beta-no-safety-score-required/}
\BIBentrySTDinterwordspacing

\bibitem{SinhaSharmaEtAl2022}
\BIBentryALTinterwordspacing
R.~Sinha, S.~Sharma, S.~Banerjee, T.~Lew, R.~Luo, S.~M. Richards, Y.~Sun, E.~Schmerling, and M.~Pavone, ``A system-level view on out-of-distribution data in robotics,'' 2022. [Online]. Available: \url{https://arxiv.org/abs/2212.14020}
\BIBentrySTDinterwordspacing

\bibitem{ElhafsiSinhaEtAl2023}
\BIBentryALTinterwordspacing
A.~Elhafsi, R.~Sinha, C.~Agia, E.~Schmerling, I.~A.~D. Nesnas, and M.~Pavone, ``Semantic anomaly detection with large language models,'' \emph{{Autonomous Robots}}, vol.~47, no.~8, pp. 1035--1055, Oct. 2023. [Online]. Available: \url{https://arxiv.org/abs/2305.11307}
\BIBentrySTDinterwordspacing

\bibitem{robitzski2021trafficlight}
\BIBentryALTinterwordspacing
D.~Robitzski, ``Watch tesla autopilot get bamboozled by a truck hauling traffic lights,'' June 2021, accessed: 2025-04-07. [Online]. Available: \url{https://futurism.com/the-byte/tesla-autopilot-bamboozled-truck-traffic-lights}
\BIBentrySTDinterwordspacing

\bibitem{robitzski2021stopsign}
\BIBentryALTinterwordspacing
------. (2021, April) Tesla keeps "slamming on the brakes" when it sees stop sign on billboard. Accessed: 2025-04-16. [Online]. Available: \url{https://futurism.com/the-byte/tesla-slamming-brakes-sees-stop-sign-billboard}
\BIBentrySTDinterwordspacing

\bibitem{SinhaElhafsiEtAl2024}
\BIBentryALTinterwordspacing
R.~Sinha, A.~Elhafsi, C.~Agia, M.~Foutter, E.~Schmerling, and M.~Pavone, ``Real-time anomaly detection and planning with large language models,'' in \emph{{Robotics: Science and Systems}}, Delft, Netherlands, Jul. 2024. [Online]. Available: \url{https://arxiv.org/abs/2407.08735}
\BIBentrySTDinterwordspacing

\bibitem{clip_2021}
\BIBentryALTinterwordspacing
A.~Radford, J.~W. Kim, C.~Hallacy, A.~Ramesh, G.~Goh, S.~Agarwal, G.~Sastry, A.~Askell, P.~Mishkin, J.~Clark, G.~Krueger, and I.~Sutskever, ``Learning transferable visual models from natural language supervision,'' in \emph{Proceedings of the 38th International Conference on Machine Learning}, ser. Proceedings of Machine Learning Research, M.~Meila and T.~Zhang, Eds., vol. 139.\hskip 1em plus 0.5em minus 0.4em\relax PMLR, 18--24 Jul 2021, pp. 8748--8763. [Online]. Available: \url{https://proceedings.mlr.press/v139/radford21a.html}
\BIBentrySTDinterwordspacing

\bibitem{mae_2022}
K.~He, X.~Chen, S.~Xie, Y.~Li, P.~Dollár, and R.~Girshick, ``Masked autoencoders are scalable vision learners,'' in \emph{2022 IEEE/CVF Conference on Computer Vision and Pattern Recognition (CVPR)}, 2022, pp. 15\,979--15\,988.

\bibitem{oquab2023dinov2}
M.~Oquab, T.~Darcet, T.~Moutakanni, H.~V. Vo, M.~Szafraniec, V.~Khalidov, P.~Fernandez, D.~Haziza, F.~Massa, A.~El-Nouby, R.~Howes, P.-Y. Huang, H.~Xu, V.~Sharma, S.-W. Li, W.~Galuba, M.~Rabbat, M.~Assran, N.~Ballas, G.~Synnaeve, I.~Misra, H.~Jegou, J.~Mairal, P.~Labatut, A.~Joulin, and P.~Bojanowski, ``Dinov2: Learning robust visual features without supervision,'' 2023.

\bibitem{hallucination_survey_2025}
\BIBentryALTinterwordspacing
L.~Huang, W.~Yu, W.~Ma, W.~Zhong, Z.~Feng, H.~Wang, Q.~Chen, W.~Peng, X.~Feng, B.~Qin, and T.~Liu, ``A survey on hallucination in large language models: Principles, taxonomy, challenges, and open questions,'' \emph{ACM Trans. Inf. Syst.}, vol.~43, no.~2, Jan. 2025. [Online]. Available: \url{https://doi.org/10.1145/3703155}
\BIBentrySTDinterwordspacing

\bibitem{kelly2017stopSign}
\BIBentryALTinterwordspacing
B.~R. Kelly, ``It's not just a stop sign,'' \emph{Kentucky Teacher}, October 2017, accessed: 2025-04-16. [Online]. Available: \url{https://www.kentuckyteacher.org/features/2017/10/its-not-just-a-stop-sign/}
\BIBentrySTDinterwordspacing

\bibitem{bommasani2021a}
R.~Bommasani, D.~A. Hudson, ..., D.~Demszky, ..., and P.~Liang, ``On the opportunities and risks of foundation models,'' \emph{arXiv preprint arXiv:2108.07258}, Jul. 2021.

\bibitem{radford2021clip}
\BIBentryALTinterwordspacing
A.~Radford, J.~W. Kim, C.~Hallacy, A.~Ramesh, G.~Goh, S.~Agarwal, G.~Sastry, A.~Askell, P.~Mishkin, J.~Clark, G.~Krueger, and I.~Sutskever, ``Learning transferable visual models from natural language supervision,'' in \emph{Proceedings of the 38th International Conference on Machine Learning}, ser. Proceedings of Machine Learning Research, M.~Meila and T.~Zhang, Eds., vol. 139.\hskip 1em plus 0.5em minus 0.4em\relax PMLR, 18--24 Jul 2021, pp. 8748--8763. [Online]. Available: \url{https://proceedings.mlr.press/v139/radford21a.html}
\BIBentrySTDinterwordspacing

\bibitem{caron2021dino}
M.~Caron, H.~Touvron, I.~Misra, H.~J\'egou, J.~Mairal, P.~Bojanowski, and A.~Joulin, ``Emerging properties in self-supervised vision transformers,'' in \emph{Proceedings of the International Conference on Computer Vision (ICCV)}, 2021.

\bibitem{darcet2023vitneedreg}
T.~Darcet, M.~Oquab, J.~Mairal, and P.~Bojanowski, ``Vision transformers need registers,'' 2023.

\bibitem{kirillov2023segmentanything}
A.~Kirillov, E.~Mintun, N.~Ravi, H.~Mao, C.~Rolland, L.~Gustafson, T.~Xiao, S.~Whitehead, A.~C. Berg, W.-Y. Lo, P.~Doll{\'a}r, and R.~Girshick, ``Segment anything,'' \emph{arXiv:2304.02643}, 2023.

\bibitem{ravi2025sam}
\BIBentryALTinterwordspacing
N.~Ravi, V.~Gabeur, Y.-T. Hu, R.~Hu, C.~Ryali, T.~Ma, H.~Khedr, R.~R{\"a}dle, C.~Rolland, L.~Gustafson, E.~Mintun, J.~Pan, K.~V. Alwala, N.~Carion, C.-Y. Wu, R.~Girshick, P.~Dollar, and C.~Feichtenhofer, ``{SAM} 2: Segment anything in images and videos,'' in \emph{The Thirteenth International Conference on Learning Representations}, 2025. [Online]. Available: \url{https://openreview.net/forum?id=Ha6RTeWMd0}
\BIBentrySTDinterwordspacing

\bibitem{minderer2022owl}
M.~Minderer, A.~Gritsenko, A.~Stone, M.~Neumann, D.~Weissenborn, A.~Dosovitskiy, A.~Mahendran, A.~Arnab, M.~Dehghani, Z.~Shen, X.~Wang, X.~Zhai, T.~Kipf, and N.~Houlsby, ``Simple open-vocabulary object detection,'' in \emph{Computer Vision – {ECCV} 2022}, S.~Avidan, G.~Brostow, M.~Cissé, G.~M. Farinella, and T.~Hassner, Eds.\hskip 1em plus 0.5em minus 0.4em\relax Springer Nature Switzerland, pp. 728--755.

\bibitem{minderer2023scaling}
\BIBentryALTinterwordspacing
M.~Minderer, A.~A. Gritsenko, and N.~Houlsby, ``Scaling open-vocabulary object detection,'' in \emph{Thirty-seventh Conference on Neural Information Processing Systems}, 2023. [Online]. Available: \url{https://openreview.net/forum?id=mQPNcBWjGc}
\BIBentrySTDinterwordspacing

\bibitem{sam2_OOD_2024}
W.~Zhao, J.~Li, X.~Dong, Y.~Xiang, and Y.~Guo, ``Segment every out-of-distribution object,'' in \emph{2024 IEEE/CVF Conference on Computer Vision and Pattern Recognition (CVPR)}, 2024, pp. 3910--3920.

\bibitem{zhou2023anomalyclip}
Q.~Zhou, G.~Pang, Y.~Tian, S.~He, and J.~Chen, ``Anomalyclip: Object-agnostic prompt learning for zero-shot anomaly detection,'' in \emph{The Twelfth International Conference on Learning Representations}, 2023.

\end{thebibliography}

\clearpage
\section*{APPENDIX}

This appendix provides additional data to support the paper's findings, as well as visualizations to improve understanding.

\subsection{Examples images of the different anomaly scenarios}
\label{sec:example_images_scenarios}

\begin{figure}[h]
    \centering
    
    \subcaptionbox{OOD Object Example 1\label{fig:ood1}}[0.48\linewidth]{%
        \includegraphics[width=\linewidth]{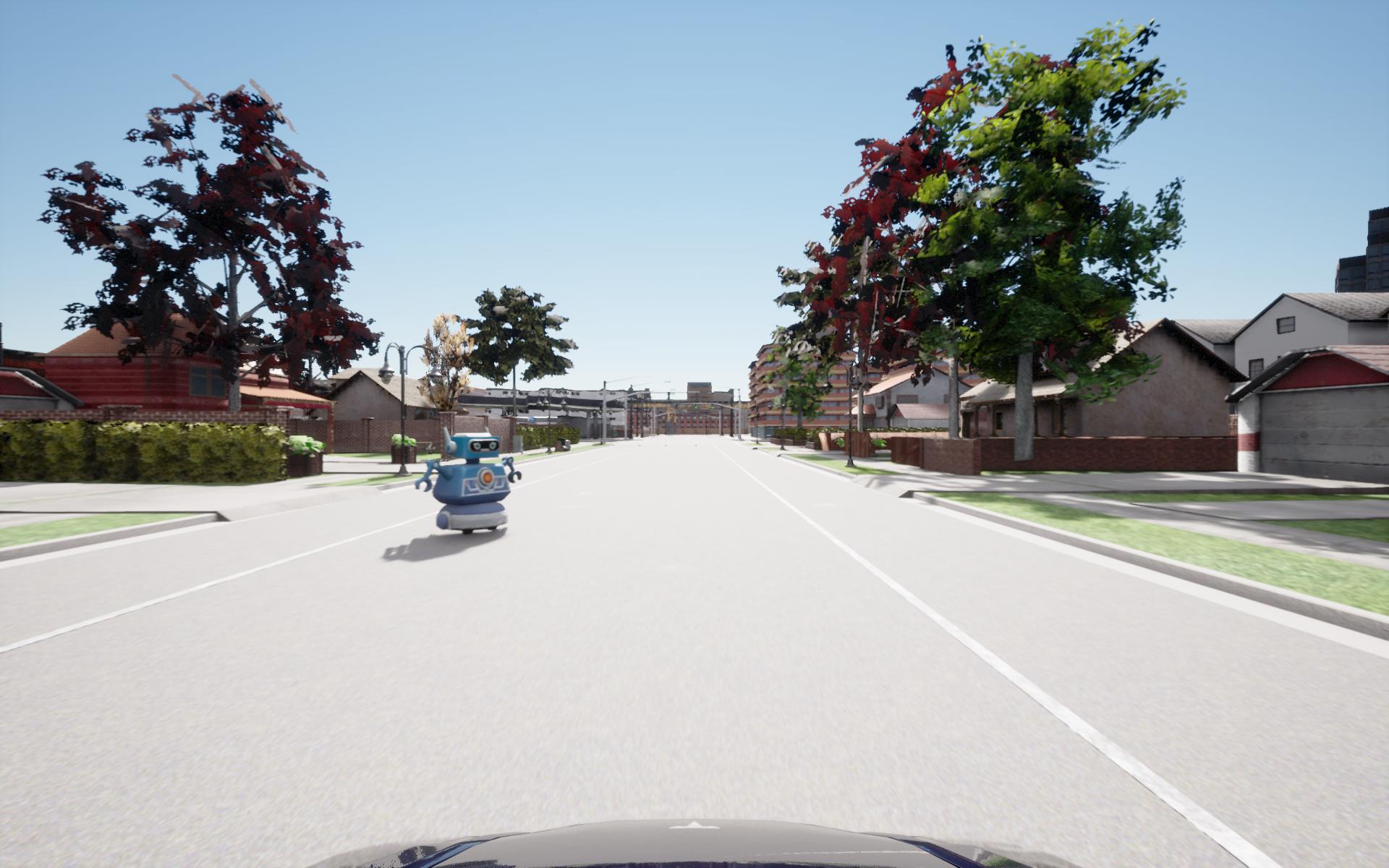}}
    \hfill
    \subcaptionbox{OOD Object Example 2\label{fig:ood2}}[0.48\linewidth]{%
        \includegraphics[width=\linewidth]{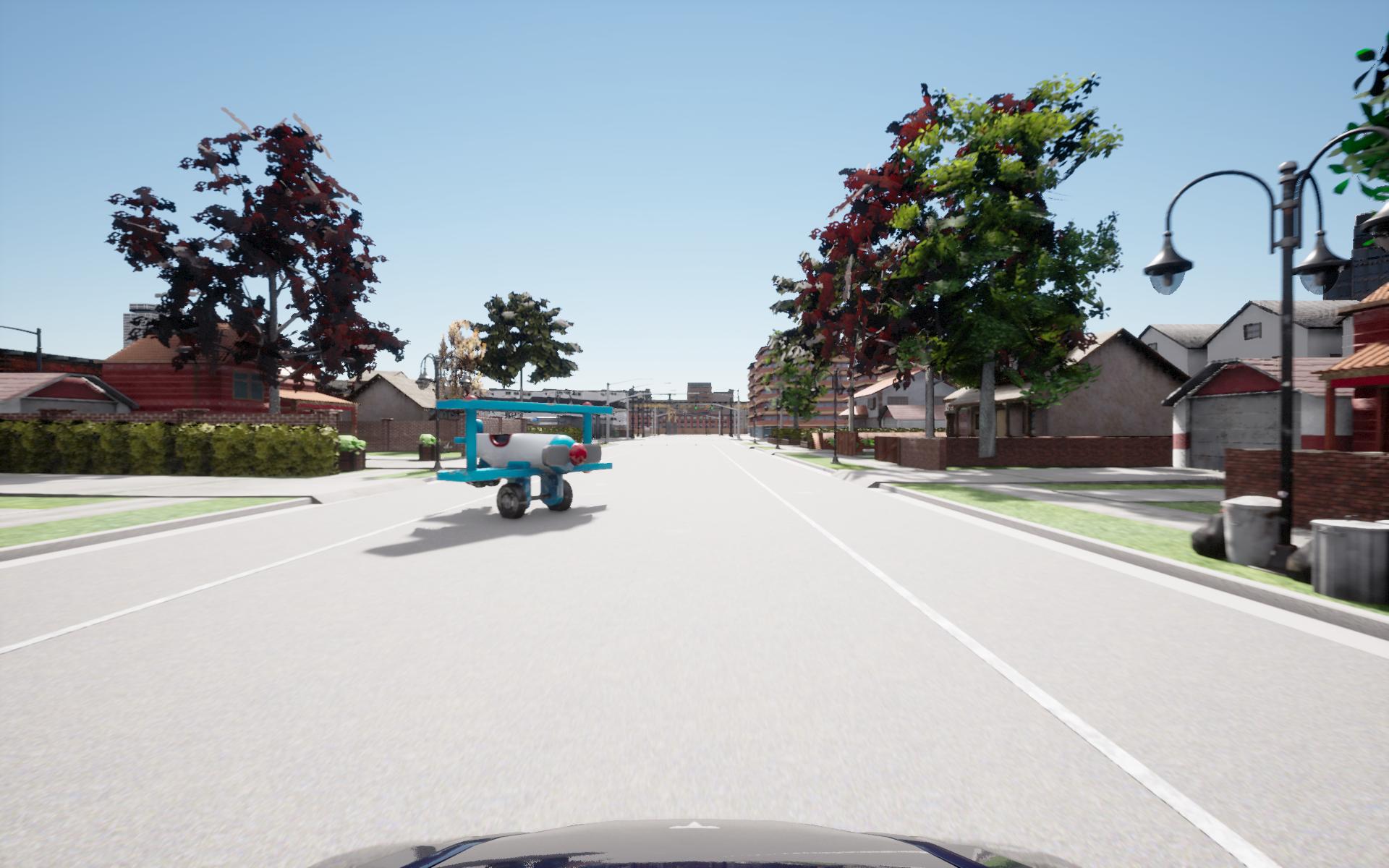}}

    \vspace{1em}

    \subcaptionbox{Traffic Light Scenario 1\label{fig:traffic1}}[0.48\linewidth]{%
        \includegraphics[width=\linewidth]{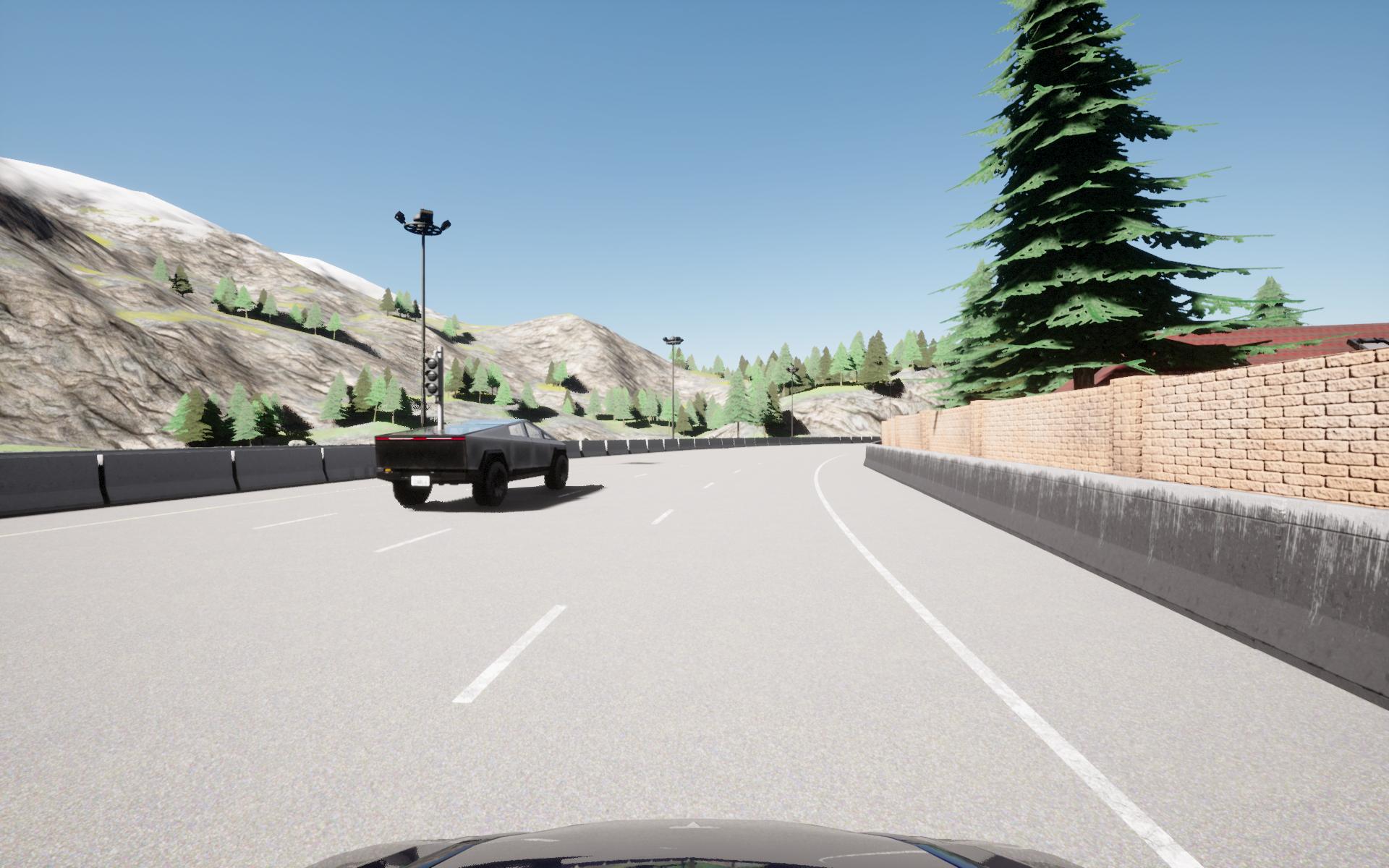}}
    \hfill
    \subcaptionbox{Traffic Light Scenario 2\label{fig:traffic2}}[0.48\linewidth]{%
        \includegraphics[width=\linewidth]{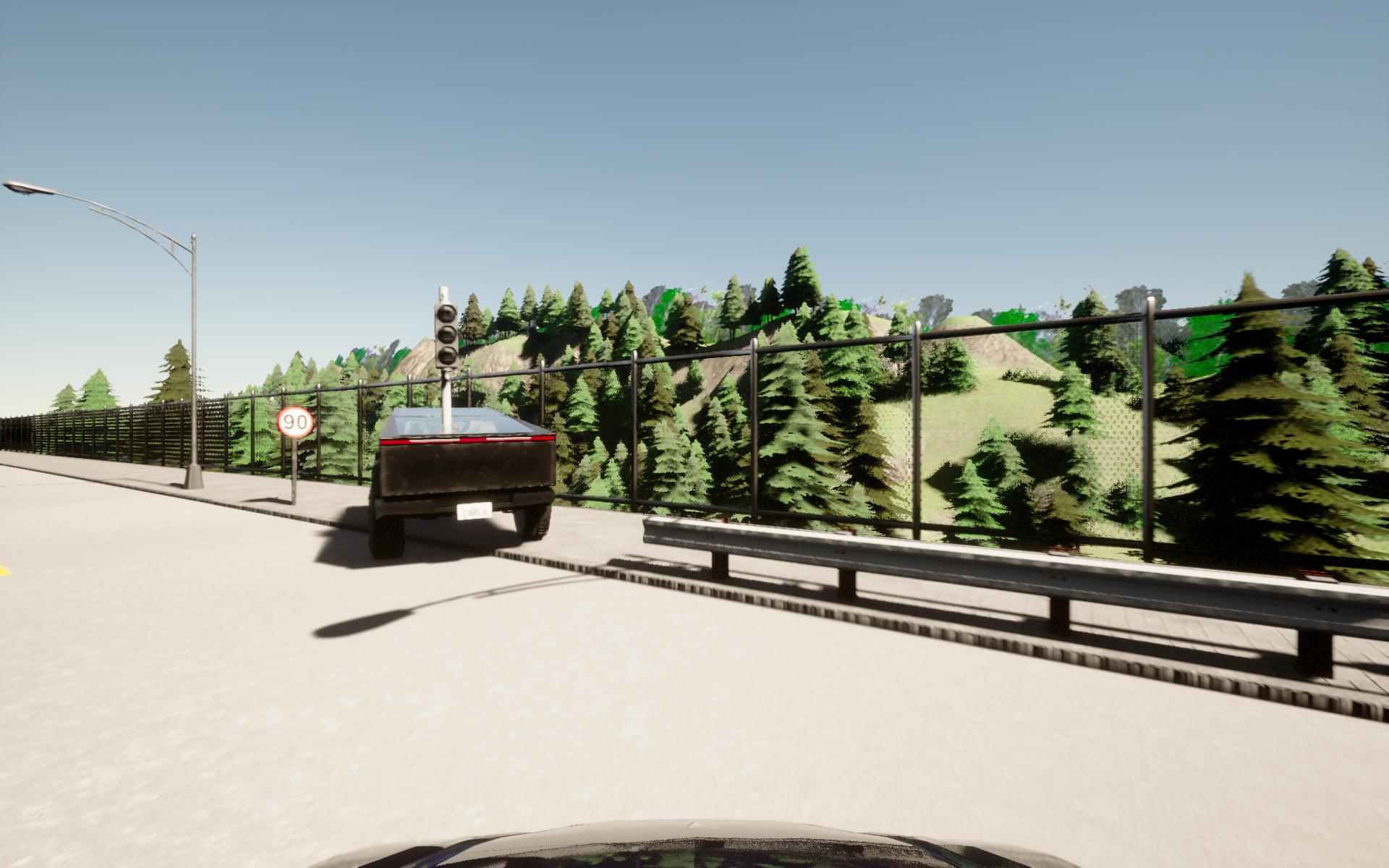}}

    \vspace{1em}

    \subcaptionbox{Stop Sign Scenario 1\label{fig:stop1}}[0.48\linewidth]{%
        \includegraphics[width=\linewidth]{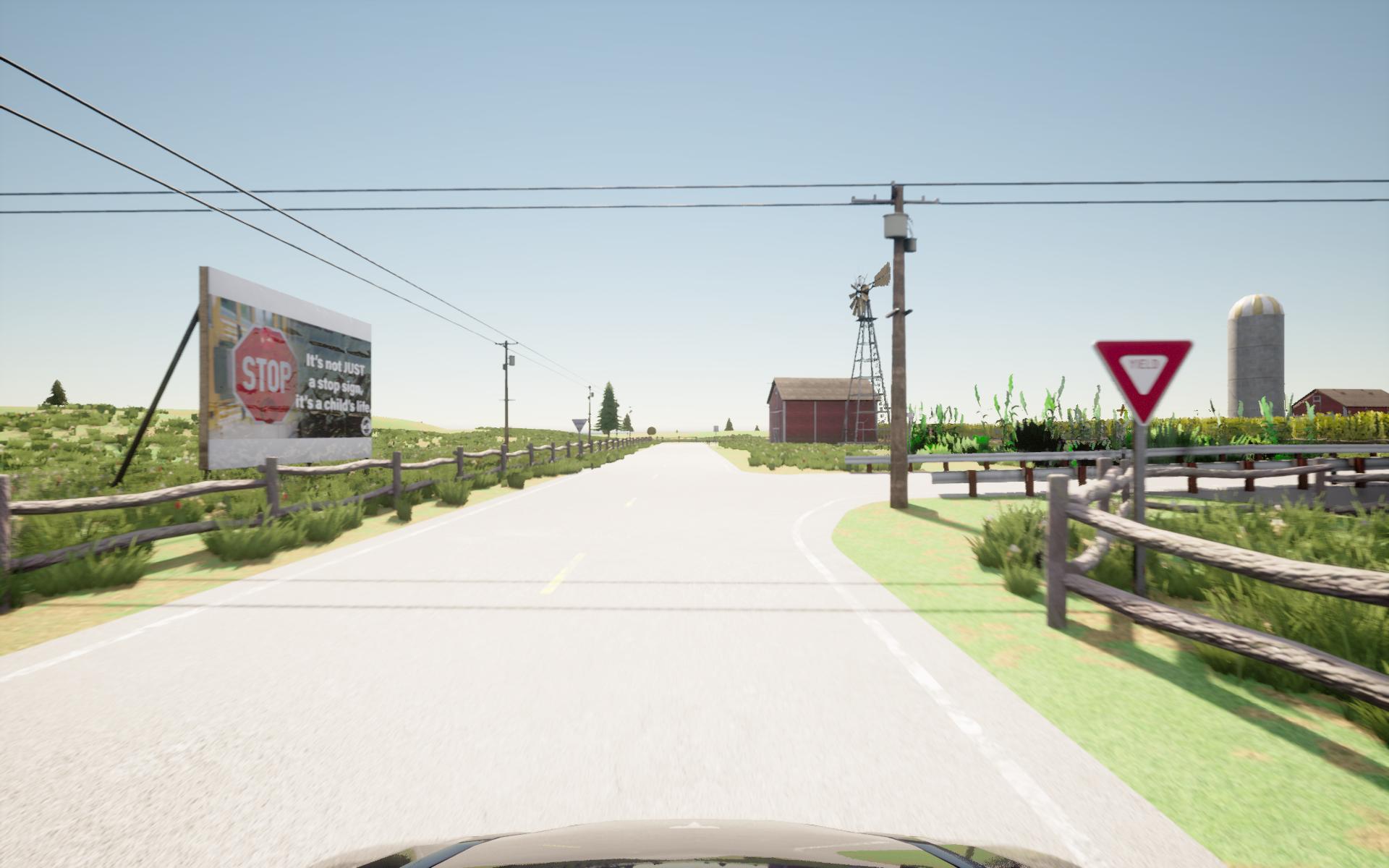}}
    \hfill
    \subcaptionbox{Stop Sign Scenario 2\label{fig:stop2}}[0.48\linewidth]{%
        \includegraphics[width=\linewidth]{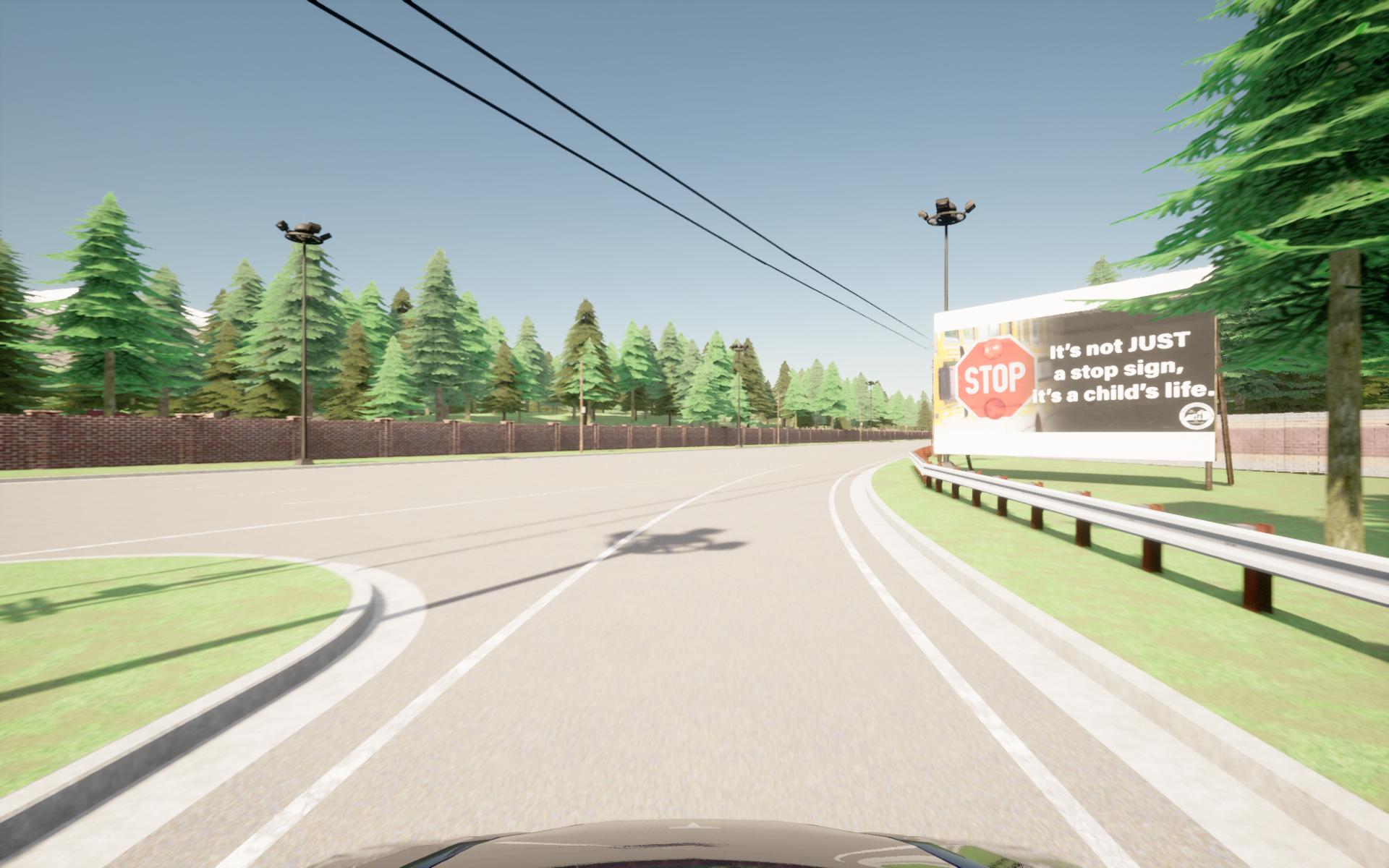}}

    \caption{Example images from the different scenarios used for evaluation: OOD object (top), traffic light (middle), and stop sign (bottom).}
    \label{fig:scenario_examples}
\end{figure}

Figure~\ref{fig:scenario_examples} shows example images from the evaluated scenarios: OOD objects, traffic lights, and stop signs.

\subsection{Processing Steps Visualization}
The following section visualizes key steps of the anomaly detection pipelines to illustrate intermediate results and support interpretation.

\subsubsection{Embedding-Based Anomaly Detection Visualization}
\begin{figure*}[t]
    \centering
    \begin{subfigure}[b]{0.45\textwidth}
        \centering
        \includegraphics[width=\textwidth]{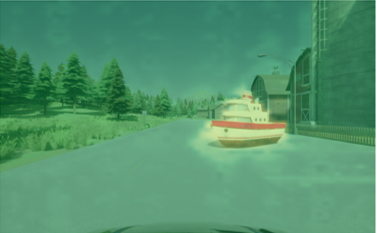}
        \caption{Anomaly scoring heatmap}
        \label{fig:scoring}
    \end{subfigure}
    \hfill
    \begin{subfigure}[b]{0.45\textwidth}
        \centering
        \includegraphics[width=\textwidth]{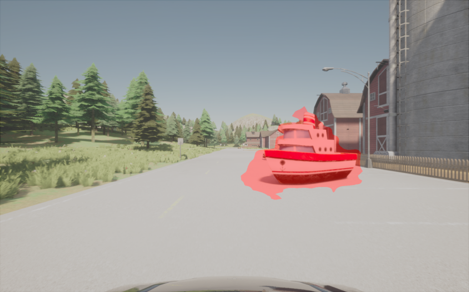}
        \caption{Anomaly classification result.}
        \label{fig:classification}
    \end{subfigure}
    \caption{\textbf{Embedding-based anomaly detection pipeline:} (a) computes anomaly scores from embedding distances. (b) shows the final classification after thresholding.}
    \label{fig:embedding_pipeline}
\end{figure*}

To better understand the pipeline, Fig.~\ref{fig:embedding_pipeline} visualizes intermediate steps of the embedding-based anomaly detection. The left image shows the anomaly score heatmap, where higher values (yellow) indicate stronger anomalies. As seen in the score histograms (Fig.~\ref{fig:score_distribution}), the scores do not reach the maximum dissimilarity of 1, yet the anomalous object is clearly visible. Nominal regions, such as the surrounding vegetation, remain consistently green, indicating low anomaly scores. The right image shows the final classification after thresholding, highlighting the detected anomaly (red areas). The mask appears slightly blurry and extends beyond the actual anomalous object. This is likely due to the limited number of embeddings (256) used for the entire image, resulting in reduced spatial resolution.

\subsubsection{Instance-Based Anomaly Detection Visualization}

\begin{figure*}[t]
    \centering
    \begin{subfigure}[b]{0.24\textwidth}
        \centering
        \includegraphics[width=\textwidth]{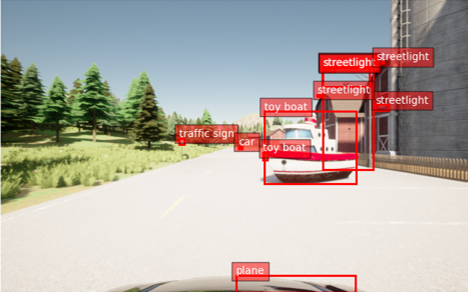}
        \caption{Object detection}
    \end{subfigure}
    \hfill
    \begin{subfigure}[b]{0.24\textwidth}
        \centering
        \includegraphics[width=\textwidth]{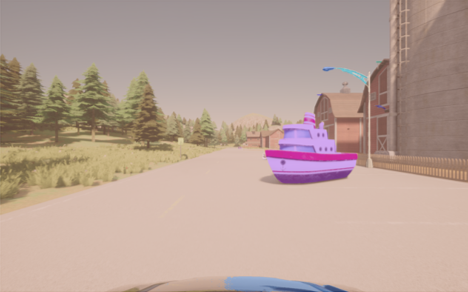}
        \caption{Instance segmentation..}
    \end{subfigure}
    \hfill
    \begin{subfigure}[b]{0.24\textwidth}
        \centering
        \includegraphics[width=\textwidth]{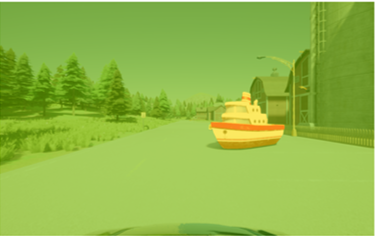}
        \caption{Anomaly scoring heatmap}
    \end{subfigure}
    \hfill
    \begin{subfigure}[b]{0.24\textwidth}
        \centering
        \includegraphics[width=\textwidth]{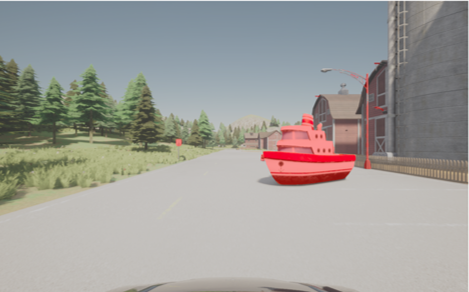}
        \caption{Anomaly classification result}
    \end{subfigure}
    \caption{\textbf{Instance-based anomaly detection pipeline:} (a) shows the object detection output. (b) visualizes the segmentation mask of SAM2 that used the bounding boxes for prompting. (c) illustrates the anomaly score heatmap. (d) presents the final classification output, highlighting anomalous objects. This visualization helps to better understand the detection steps.}

    \label{fig:instance_pipeline}
\end{figure*}

To better understand the instance-based pipeline, Fig.~\ref{fig:instance_pipeline} shows all intermediate steps. The first image presents object detection. While the boat is correctly detected, the streetlight is detected multiple times. The second image shows SAM2 segmentation, which performs well on the boat but struggles with the streetlight due to overlapping boxes. The background is grouped into a single mask.

The third image displays the anomaly score heatmap. The contrast between nominal and anomalous objects is weaker than in the embedding-based version—all areas appear mostly yellow—but the boat remains distinguishable. This is likely due to embedding averaging within the large background mask. The final image shows the classification result, highlighting the anomalous boat with sharp boundaries. Minor false positives appear at the traffic sign and streetlight, likely due to imperfect detection and segmentation, resulting in partial or unseen objects and embeddings.

\subsection{Prompt Used for ChatGPT-4o}
\label{sec:gpt4_prompt}
Below is the prompt used for semantic anomaly detection with GPT-4o.

\begin{tcolorbox}[colback=gray!5!white,colframe=gray!75!black,title=ChatGPT-4o Prompt, breakable]
I am the fault monitor for a vision-based autonomous vehicle. My task is to critically evaluate the vehicle’s observations in a given image and determine whether they contain any elements that could lead to unsafe, unpredictable, or traffic rule-violating actions.

For each object or element in the scene, I will analyze and answer the following:
\begin{enumerate}
  \item Is this common to see while driving?
  \item Can this influence the vehicle’s behavior?
  \item Can the vehicle drive safely in its presence?
  \item Can this cause the vehicle to make unpredictable or unsafe maneuvers?
\end{enumerate}

Based on the analysis, I will classify each object as either:
\begin{itemize}
  \item \textbf{Nominal (Normal) Observation}: Does not impact safety or traffic rule compliance.
  \item \textbf{Anomaly}: May pose risks, lead to unsafe behavior, or cause violations.
\end{itemize}

I will also evaluate the entire scene and classify it as either \textbf{Nominal} or an \textbf{Anomaly}, with reasoning.

\textbf{Output Format:}
\begin{enumerate}
  \item \textbf{Individual Object Assessments:} Answer the four questions and classify each object.
  \item \textbf{Overall Scenario Analysis:} Reasoning based on all objects.
  \item \textbf{Overall Scenario Classification:} Nominal or Anomaly.
\end{enumerate}

\textbf{Example:}
The vehicle observes:
\begin{itemize}
  \item A cyclist riding on the sidewalk \\
  \textit{Classification: Normal}
  \item A car parked in a no-parking zone \\
  \textit{Classification: Normal}
  \item A pedestrian carrying a bright green balloon \\
  \textit{Classification: Anomaly}
\end{itemize}

\textbf{Overall Scenario Classification:} Anomaly.\\

For the given image execute the task as described in the example and strictly follow and adhere to the provided output structure. \\

I am driving on the road and I see:

\end{tcolorbox}

\subsection{GPT-4o failures}
\label{sec:gpt4o-failures}
\begin{figure}[h]
    \centering

    % Failure Case 1
    \subcaptionbox{GPT-4o Failure Case 1\label{fig:gpt4o_fail1}}[0.48\linewidth]{%
        \includegraphics[width=\linewidth]{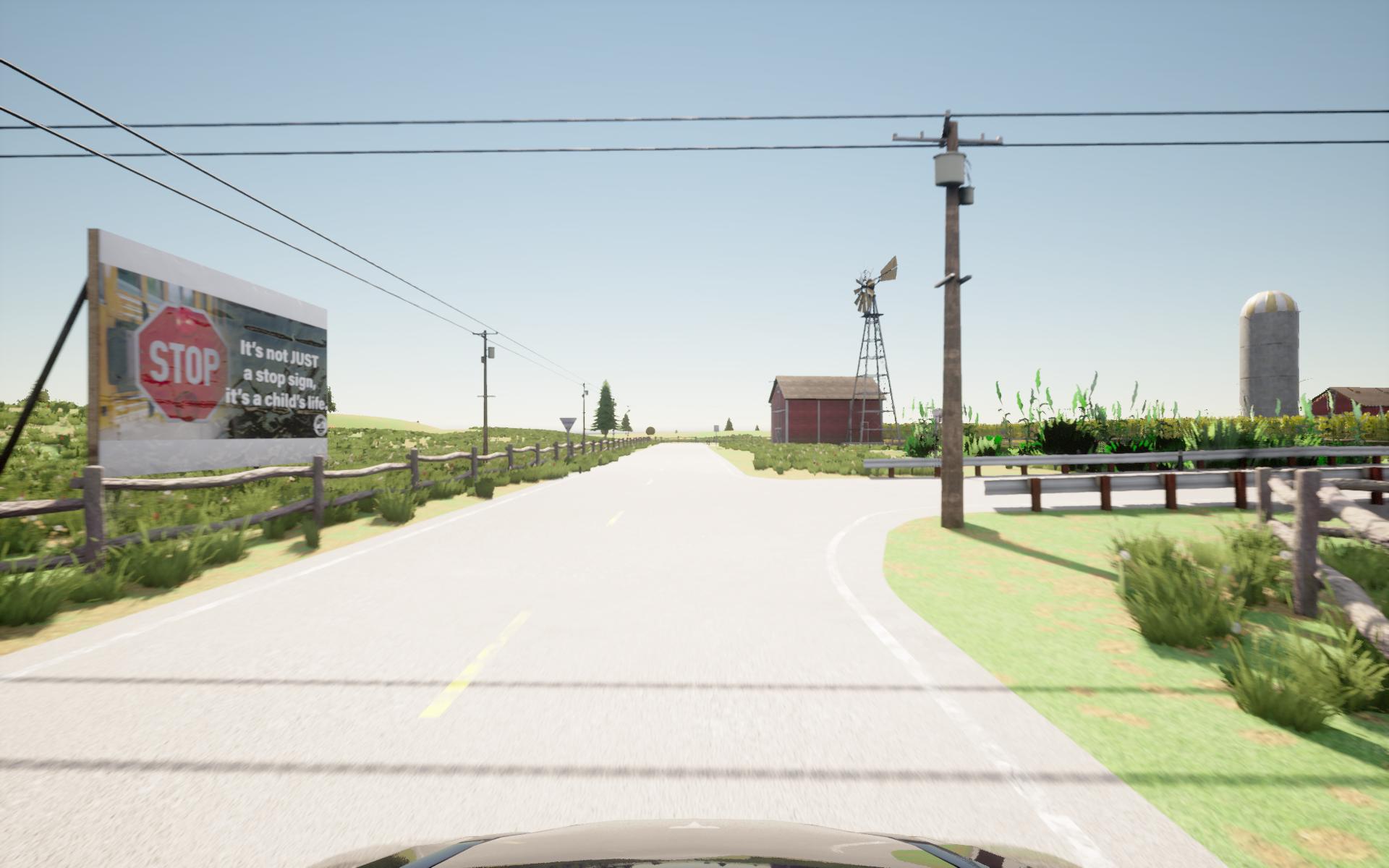}}
    \hfill
    \subcaptionbox{GPT-4o Response 1\label{fig:gpt4o_text1}}[0.48\linewidth]{%
        \scriptsize
        \begin{minipage}[t]{\linewidth}
        \textbf{Overall Scenario Classification:} Nominal \\
        "...Billboard with stop sign message: Nominal. The billboard provides an important safety reminder about stopping, although it does not directly impact vehicle operation."
        \end{minipage}}

    \vspace{1em}

    % Failure Case 2
    \subcaptionbox{GPT-4o Failure Case 2\label{fig:gpt4o_fail2}}[0.48\linewidth]{%
        \includegraphics[width=\linewidth]{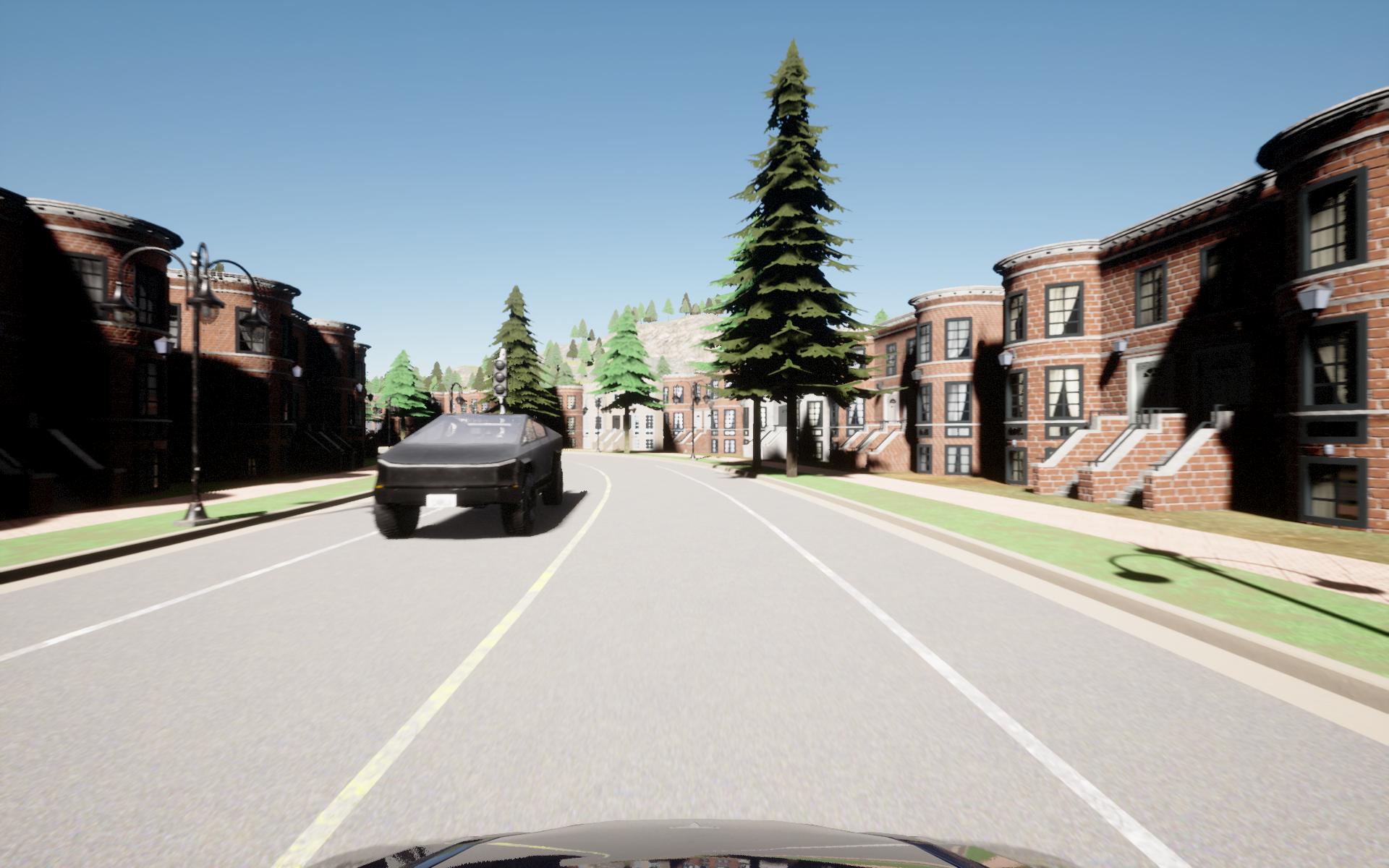}}
    \hfill
    \subcaptionbox{GPT-4o Response 2\label{fig:gpt4o_text2}}[0.48\linewidth]{%
        \scriptsize
        \begin{minipage}[t]{\linewidth}
        \textbf{Overall Scenario Classification:} Anomaly \\
        "...the improperly aligned vehicle could pose a risk... The presence of traffic lights ahead is nominal, but it is unclear from the image if the light is red, yellow, or green..."
        \end{minipage}}

    \vspace{1em}

    % Failure Case 3
    \subcaptionbox{GPT-4o Failure Case 3\label{fig:gpt4o_fail3}}[0.48\linewidth]{%
        \includegraphics[width=\linewidth]{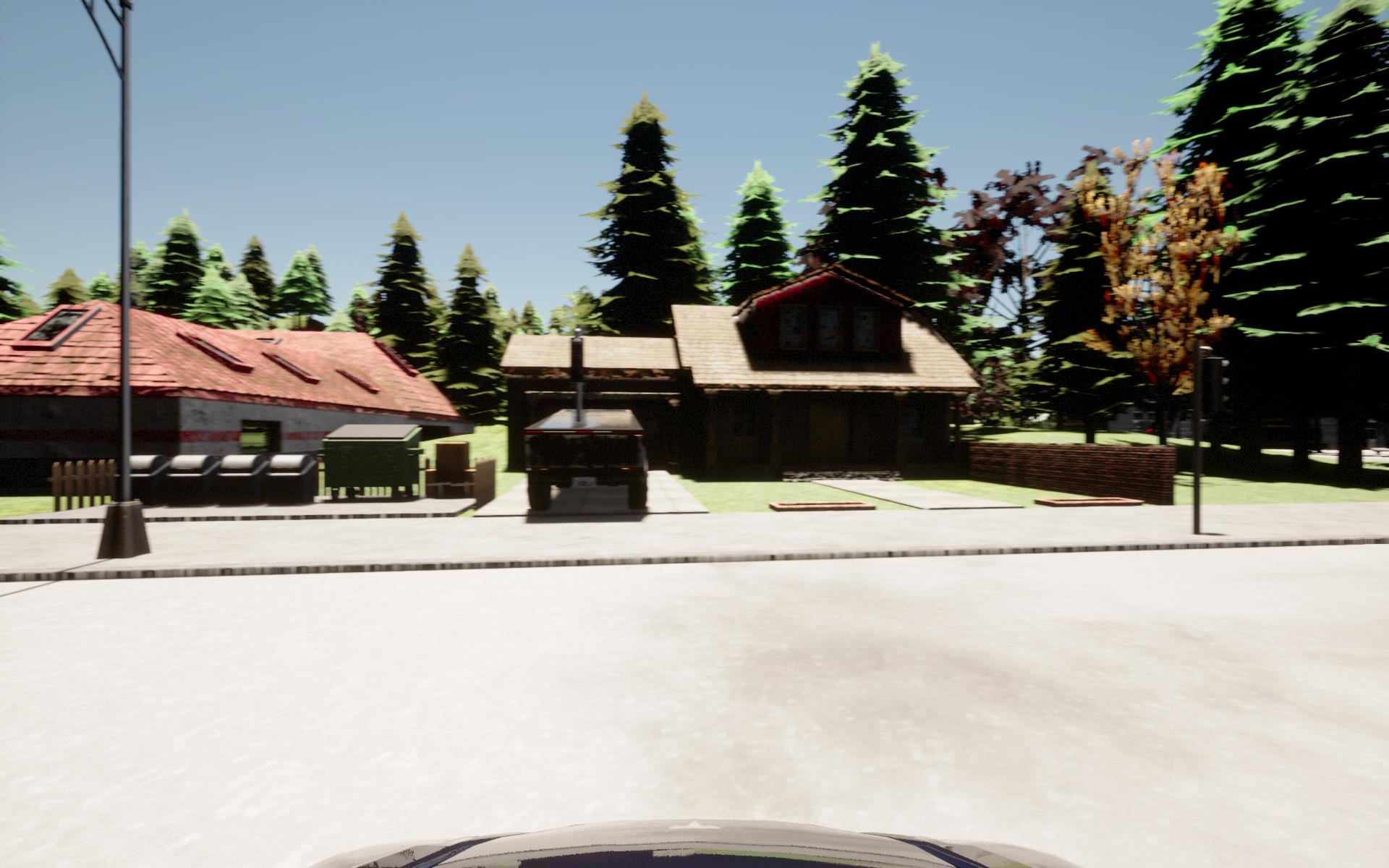}}
    \hfill
    \subcaptionbox{GPT-4o Response 3\label{fig:gpt4o_text3}}[0.48\linewidth]{%
        \scriptsize
        \begin{minipage}[t]{\linewidth}
        \textbf{Overall Scenario Classification:} Nominal \\
        "...Parked vehicle in driveway: Nominal observation. Correctly parked off the main road does not interfere with traffic or pedestrian paths"
        \end{minipage}}

    \caption{GPT-4o failure cases where anomalies are either misclassified as nominal or flagged for incorrect reasons. The responses illustrate reasoning behind the misclassification.}
    \label{fig:gpt4o_failures}
\end{figure}

As mentioned in Section~\ref{sec:result_analysis}, GPT-4o struggles with more subtle semantic anomalies. This seems partly due to a limited understanding of what defines a semantic anomaly. GPT-4o sometimes classifies semantic anomalies as nominal, or it detects them but for the wrong reason. Examples are shown in Fig.~\ref{fig:gpt4o_failures}.

In Example~\ref{fig:gpt4o_fail1}, the stop sign on the billboard is classified as nominal. GPT-4o explains that it provides an important safety reminder, but does not consider it a potential risk (\ref{fig:gpt4o_text1}). While the reasoning itself is valid, it may overlook the relevance of such cues in the context of autonomous driving.

In Example~\ref{fig:gpt4o_fail2}, the image is correctly classified as an anomaly, but the reason given is the vehicle’s misalignment, not the fact that it carries a traffic light. The traffic light is mentioned, but only to note that its color cannot be determined (\ref{fig:gpt4o_text2}).

These two cases show that GPT-4o may lack the domain-specific context needed to correctly identify and interpret semantic anomalies. In Example~\ref{fig:gpt4o_fail3}, the traffic light is not mentioned at all, and the truck is classified as nominal (\ref{fig:gpt4o_text3}). This could be due to a domain shift, as GPT-4o was likely trained mostly on real-world images. These issues highlight the limitations of using GPT-4o in a zero-shot setting. Fine-tuning on driving-specific tasks or providing more targeted in-context examples could help mitigate these problems.

\subsection{Examples of failures in the object detector}
\label{sec:object_detector_segmentation}

\begin{figure}[h]
    \centering
    \includegraphics[width=0.95\linewidth]{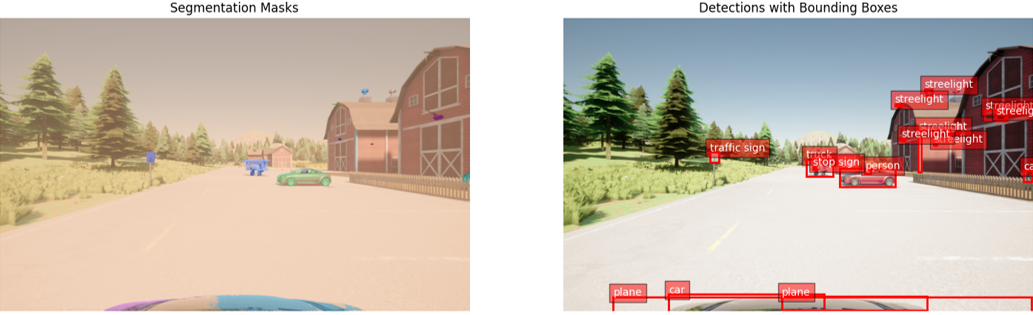}

    \vspace{1em}

    \includegraphics[width=0.95\linewidth]{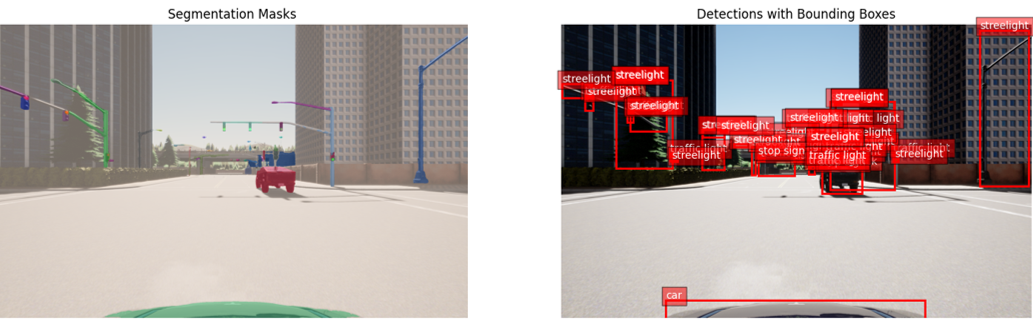}

    \vspace{1em}

    \includegraphics[width=0.95\linewidth]{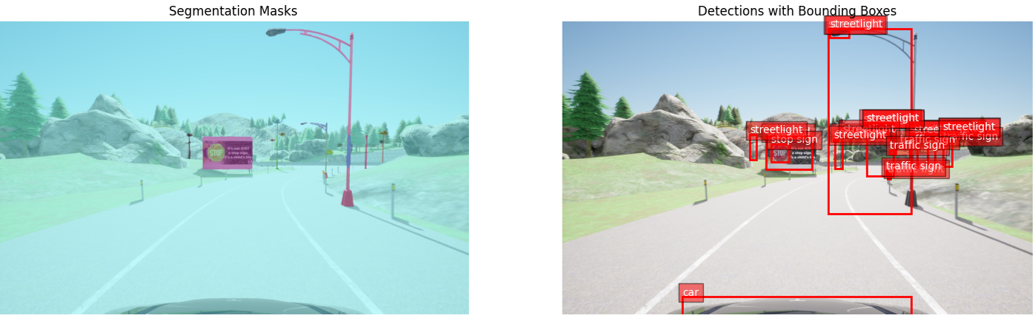}

    \caption{Examples where false positives in the object detector lead to oversegmentation of certain objects. Each row shows the predicted segmentation masks (left) and the corresponding detections with bounding boxes (right).}
    \label{fig:oversegmentation_examples}
\end{figure}

For instance segmentation, a combination of OWLv2 and SAM2 is used, where bounding boxes from OWLv2 are used to prompt SAM2. However, OWLv2 often produces false positives on synthetic CARLA images, especially for objects like traffic signs and street lights. Examples are shown in Fig.~\ref{fig:oversegmentation_examples}. Similar issues with OWLv2 have been reported in \cite{SinhaElhafsiEtAl2024}. Multiple overlapping detections can lead SAM2 to oversegment objects, splitting them into fragments. These fragments may be misclassified as anomalies due to their novel or inconsistent appearance relative to the nominal dataset. Finetuning the object detector on the specific dataset or using real-world images instead of synthetic ones could help reduce these issues.

% \begin{figure}[h]
%     \centering
%     \subcaptionbox{Example 1\label{fig:overseg1}}[0.95\linewidth]{%
%         \includegraphics[width=\linewidth]{Pictures/over_undersegmentation_01.png}}

%     \vspace{1em}

%     \subcaptionbox{Example 2\label{fig:overseg2}}[0.95\linewidth]{%
%         \includegraphics[width=\linewidth]{Pictures/over_undersegmentation_02.png}}

%     \vspace{1em}

%     \subcaptionbox{Example 3\label{fig:overseg3}}[0.95\linewidth]{%
%         \includegraphics[width=\linewidth]{Pictures/over_undersegmentation_03.png}}

%     \caption{Examples where false positives in the object detector lead to oversegmentation of certain objects. Each row shows the predicted segmentation masks (left) and the corresponding detections with bounding boxes (right).}
%     \label{fig:oversegmentation_examples}
% \end{figure}

% \subsection{Dataset size impact}

% \subsection{PCA embedding dimension}

% \subsection{PCA clustering}

\end{document}